\documentclass{article}
\usepackage{amsmath}
\usepackage{amssymb}
\usepackage{multicol}
\usepackage[bookmarks=true]{hyperref}
\usepackage{yhmath}
\usepackage{multirow}
\usepackage{algorithm}
\usepackage{algpseudocode}
\usepackage{tikz}
\usepackage{makecell}
\algrenewcommand{\algorithmicrequire}{\textbf{Input:}}
\algrenewcommand{\algorithmicensure}{\textbf{Output:}}

\DeclareMathOperator*{\argmin}{argmin} 

\usepackage{multicol}
\usepackage{graphicx}
\usepackage{url}
\usepackage{wrapfig} 
\usepackage[bookmarks=true]{hyperref}
\usepackage{xcolor}
\usepackage{subcaption}
\usepackage{float}
\usepackage{bm}
\usepackage{caption}
\usepackage{lipsum}  
\usepackage[para,online,flushleft]{threeparttable}
\usepackage{booktabs} 
\usepackage{siunitx}
\usepackage{etoolbox} 
\usepackage[normalem]{ulem}
\usepackage{xspace}
\usepackage[inline]{enumitem}
\usepackage[preprint]{corl_2025} 

\newcommand{\citept}{\cite}
\newcommand{\parab}[1]{\noindent\textbf{#1}}

\DeclareRobustCommand\BoxedLabel[1]{\tikz[baseline=(label.base)]\node[draw,inner sep=2pt](label){#1};}

\newcommand{\algname}{KineSoft\xspace}
\title{\algname: Learning Proprioceptive Manipulation Policies with Soft Robot Hands}

\author{
  Uksang Yoo$^{1}$, 
  Jonathan Francis$^{1,2}$, 
  Jean Oh$^{1}$, 
  Jeffrey Ichnowski$^{1}$ \\
  $^{1}$Carnegie Mellon University \quad
  $^{2}$Bosch Center for AI
}

\begin{document}
\makeatletter
\maketitle

\vspace{0.3cm}
\begin{abstract}
Underactuated soft robot hands offer inherent safety and adaptability advantages over rigid systems. While imitation learning shows promise for acquiring complex dexterous manipulation skills, adapting existing methods to soft robots presents unique challenges in state representation and data collection. We propose \algname, a framework for direct kinesthetic teaching of soft robotic hands that leverages their natural compliance as a skill teaching advantage rather than only as a control challenge. 
With KineSoft, human demonstrators physically guide the robot while the system learns to associate proprioceptive patterns with successful manipulation strategies. 
\algname makes three key contributions: (1) a shape-based imitation learning framework that uses proprioceptive feedback to ground diffusion-based policies (2) a low-level shape-conditioned controller that enables precise tracking of desired shape trajectories, (3) a sim-to-real learning approach to soft robot mesh shape sensing with an internal strain-sensing array. 
In physical experiments, we demonstrate the superiority of \algname over the strain-based policy baseline in six in-hand manipulation tasks involving both rigid and deformable objects. 
\algname's results suggest that embracing the inherent properties of soft robots leads to intuitive and robust dexterous manipulation capabilities. Videos and code will be available upon final decision at \href{https://kinesoft-policy.github.io}{\texttt{kinesoft-policy.github.io}}.
\end{abstract}

\begin{center}
\vspace{-0.5cm}
\includegraphics[width=0.7\linewidth]{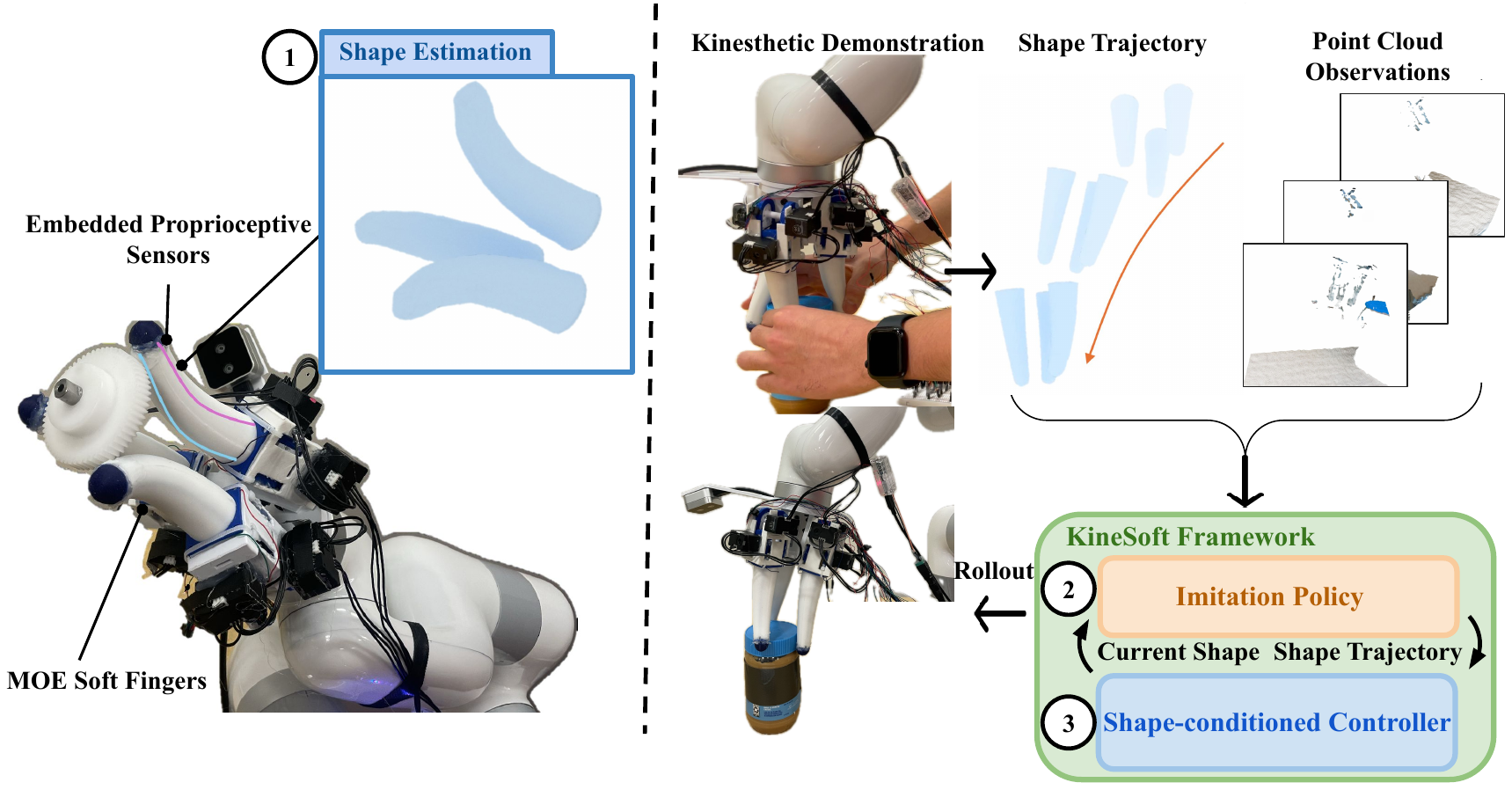}
\captionof{figure}{
  \textbf{KineSoft} is a framework for learning from kinesthetic demonstration, enabling free-shaped soft end-effectors to perform dexterous manipulation. Three key components are: 1) a proprioceptive model for high-fidelity shape estimation, 2) diffusion-based imitation learning for predicting the changes in shape and end-effector poses, and 3) a shape-conditioned controller that allows the soft hand to track given shape trajectories.
}
\label{figure:front}
\vspace{-0.3cm}
\end{center}

\section{Introduction}

Underactuated soft robot hands offer key advantages over rigid counterparts, including inherent safety through material compliance~\citep{yoo2025soft, gilday2024rigid} and robust adaptability to uncertain object geometries~\citep{bhatt2022surprisingly, homberg2019robust, yao2024soft}. These properties make them well-suited for applications requiring contact-rich and reliable interactions such as assistive care, delicate object manipulation, fruit picking, and collaborative manufacturing~\citep{yoo2025soft, firth2022anthropomorphic}. However, imparting dexterous in-hand manipulation skills to soft hands remains challenging. Existing methods for soft robot manipulation often rely on hand-crafted primitives~\citep{bhatt2022surprisingly, yao2024soft, abondance2020dexterous, jitosho2025flying} that necessitate expert operators and limit system adaptability. 

Recent advances in imitation learning, particularly frameworks like diffusion policy, have shown promise in teaching complex manipulation skills~\citep{chi2023diffusion, memmel2025strap, hu2023toward}. These approaches have been successfully applied to various tasks, from long-horizon mobile manipulation with rigid grippers~\citep{fu2024mobile} to deformable object manipulation with simple end-effectors~\citep{yoo2024ropotter}. Unlike reinforcement learning methods that depend on carefully crafted reward functions and simulation environments~\citep{qi2025simple, wang2024penspin}, imitation learning requires only demonstration trajectories of successful task execution. However, collecting such demonstrations for soft robots presents challenges: demonstration collection methods for rigid articulated robots generally do not apply to underactuated soft robots and their virtually infinite degrees of freedom.  Teleoperation interfaces~\citep{qin2023anyteleop} designed for rigid anthropomorphic hands fail to capture the unique capabilities and constraints of underactuated soft end-effectors. Soft robots often lack a direct mapping to rigid human hand joints. Additionally, standard robot state representations for rigid robots in imitation learning frameworks~\citep{ze20243d}, such as rigid transformation poses, are ill-suited to provide meaningful state information for continuously deforming structures. Despite recent advances in expressive state representation learning for soft robots~\citep{wang2023soft, yoo2023toward, yoo2024poe}, these have not yet been applied to skill learning frameworks for in-hand manipulation. These limitations have restricted the application of imitation learning to soft robotic manipulation.


We propose KineSoft, a hierarchical framework to enable direct kinesthetic teaching of soft robot hands skills. The key insight is that soft robots' natural compliance provides an advantage for teaching rather than just a control challenge. Unlike rigid robots, whose joints and linkages often resist human guidance due to strict kinematic constraints, soft hands can be easily deformed by a human demonstrator into desired poses. This allows demonstrations to be collected by physically manipulating the hand without violating mechanical limits or producing infeasible configurations. KineSoft has three key components (Figure~\ref{figure:front}). (1) The proprioceptive system achieves state-of-the-art shape estimation using internal strain-sensing arrays and a model trained on large simulated data of the robot's high-dimensional configurations. These sensors provide rich proprioceptive feedback while preserving the hand's natural compliance, allowing KineSoft to capture detailed information about the hand's deformation state during manipulation tasks in real time. (2) We train an imitation policy on these shape trajectories and use it to generate deformation trajectories during rollout. (3) KineSoft's low-level shape-conditioned controller then tracks these desired shapes. Experiments demonstrate that KineSoft achieves accurate shape state estimation, mesh-based trajectory tracking through the shape-conditioned controller, and high performance in learned manipulation skills through these shape-based representations.


In summary, this paper contributes: \begin{enumerate*}[label=\textbf{\roman*)}]
    \item KineSoft, a framework for learning from kinesthetic demonstrations for soft robot hands that enables dexterous in-hand manipulation,
    \item  a state-of-the-art proprioceptive shape estimation approach using strain sensing integrated with soft robot hands that enables precise tracking of finger deformations during contact-rich tasks,
    \item  shape-conditioned controller for tracking the generated deformation trajectories and performing dexterous manipulation tasks, and
    \item simulated dataset and trained model for state estimation and control, which we demonstrate can be readily deployed to open-source soft robot hands, such as the MOE platform~\citep{yoo2025soft}.
\end{enumerate*}


\section{Related Work}

\paragraph{Learning for in-hand dexterity.} Recent advances in reinforcement learning have driven significant progress in rigid robot in-hand dexterity~\citep{wang2024penspin, qi2025simple, yao2024soft, andrychowicz2020learning}. However, these approaches face challenges in real-world deployment due to difficulty in creating resettable training environments. To mitigate these issues, many methods transfer policies trained in simulation to the real world, which has shown success with rigid robots with fine tuning in the real world~\citep{qi2025simple, wang2024penspin} but remains less applicable for soft robots due to the complexities involved in modeling deformable materials, forward kinematics, and contact dynamics~\citep{della2023model}. Despite recent interest in leveraging reinforcement learning for soft robot arm control and trajectory tracking~\citep{thuruthel2018model, schegg2023sofagym,bhagat2019deep}, difficulty in modeling simultaneous contact and soft robot body deformation dynamics have hindered their application to soft robot dexterous manipulation.

Imitation learning has emerged as a promising alternative for reducing the reliance on explicit physics simulation environments, enabling robots to acquire manipulation skills efficiently with real-world data~\citep{johns2021coarse, chi2023diffusion, ze20243d}. In-hand manipulation skills through imitation learning are typically achievable with anthropomorphic hands that provide a direct mapping between human hand and robot poses~\citep{arunachalam2023holo, arunachalam2023dexterous, wei2024wearable}. However, despite the inherent benefits in safety and dexterity through compliance of soft robot hands, they face unique challenges due to the absence of reliable proprioceptive feedback~\citep{weinberg2024survey} and thus there is a lack of practical frameworks for collecting demonstrations for soft robots. To the best of our knowledge, \textit{KineSoft} is the first framework that effectively leverages passive compliance of soft robots to collect demonstrations and enables soft robots to acquire dexterous in-hand manipulation skills.


\paragraph{Soft robot dexterity.} Soft robot hands excel in grasping and manipulation tasks due to their material compliance, allowing for passive adaptation to diverse object geometries~\citep{rus2015design, zhou2023soft}. This compliance facilitates robust grasping and safe interactions with humans and delicate objects by distributing contact forces~\citep{abondance2020dexterous, liu2024skingrip, yoo2025soft}. Recent advances in soft robots have aimed to improve dexterity through innovative actuator designs, material improvements, and bioinspired morphologies~\citep{puhlmann2022rbo, firth2022anthropomorphic}. Despite these strides, controlling soft hands remains a significant challenge due to their complex dynamics and the high dimensionality of their state spaces~\citep{yasa2023overview}. Consequently, learning-based approaches for soft robot dexterity have primarily focused on grasping~\citep{gupta2016learning}, while the development of dexterous in-hand manipulation skills has been hindered by the lack of available demonstration methods and reliable proprioceptive feedback~\citep{weinberg2024survey}. The unique advantages of soft robot hands in robust in-hand manipulation stem from their lack of rigid skeletal structures~\citep{pagoli2021soft}. However, this absence also introduces challenges, as soft robot kinematics differ significantly from human hand motions. The \textit{KineSoft} framework bridges these gaps by integrating novel and accurate shape estimation methods with learned imitation policies, enabling efficient skill acquisition for dexterous manipulation with soft robots.

\paragraph{Soft robot state representation.} Proprioceptive shape sensing is critical for enabling robust control in soft robots, particularly for accurate shape tracking and feedback-driven control~\citep{zhou2024integrated, ku2024soft}. Existing works often use low degree-of-freedom shape representations such as constant curvature models~\citep{stella2023piecewise, yoo2021analytical} or bending angles~\citep{wall2017method}, which fail to capture the full richness of soft robot deformation states. Toward capturing these complex deformation behaviors of soft robot manipulators, recent approaches based on mechanics models have employed Cosserat rod models and high-dimensional Frenet-Serret frames, associated with the continuum cross sections~\citep{liu2021influence}. However, updating and preserving hard constraints using these approaches is computationally expensive~\citep{liu2021influence,liu2020toward}. Recent learning-based models have introduced more expressive representations of soft robot states using point clouds~\citep{wang2023soft, yoo2023toward} and meshes~\citep{yoo2024poe, tapia2020makesense}. However, these learned representations have not been connected to policy learning for dexterous manipulation tasks. Addressing this gap, \textit{KineSoft} proposes a novel framework that leverages proprioceptive sensing and learned shape representations, based on vertex displacement fields over meshes, to facilitate dexterous  manipulation skill learning in soft robots.


\begin{figure}[t]
\includegraphics[width=1.\columnwidth]{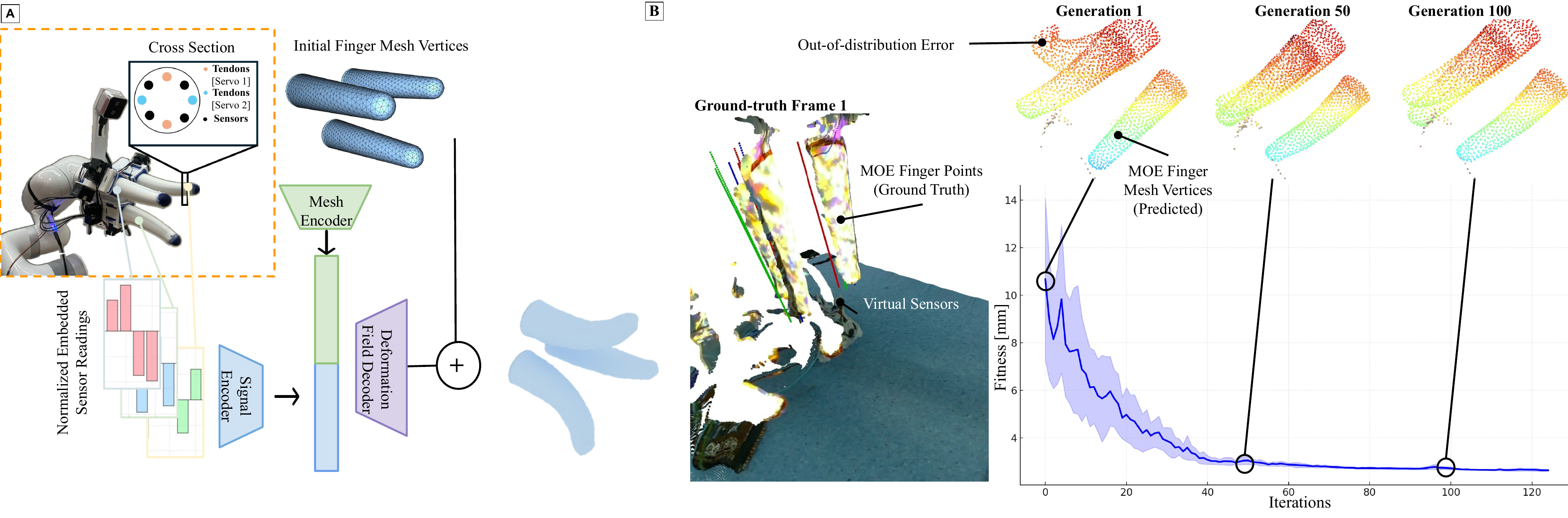}
\caption{\textbf{Proprioception network.} \BoxedLabel{A}: Network architecture for mesh shape estimation of the soft fingers. \BoxedLabel{B}: Results of domain alignment iterations where the loss converged after 200 iterations. }
\label{figure:network}
\end{figure}

\section{Problem Statement}
\label{sec:problem_statement}

We consider the problem of learning dexterous manipulation skills with a soft robot hand equipped with internal proprioceptive sensing. Let $\mathbf{M}$ denote the true continuous deformation state of the hand, which is governed by the robot’s mechanical structure and actuated through control inputs $\mathbf{U}$. Since $\mathbf{M}$ is unobservable and high- high-dimensional~\cite{katzschmann2019dynamic} in practice, we assume access to an internal sensor measurement space $\mathbf{S}$ (e.g., from strain sensors) that provides indirect information about the deformation. 

The objective is to learn a policy $\pi: \mathbf{M} \times \mathcal{O} \rightarrow \mathbf{U}$ that maps from the current deformation state and additional task-relevant observations $\mathcal{O}$ (e.g., visual input) to a control action in $\mathbf{U}$.
This setting poses three key challenges:
\begin{enumerate*}[label=\roman*)]
\item Estimating the unobservable deformation state $\mathbf{M}$ from the sensor measurement space $\mathbf{S}$;
\item Learning a reliable mapping from $\mathbf{S}$ to $\mathbf{M}$ for downstream use in control;
\item Learning a control policy $\pi$ that uses $\mathbf{M}$ and $\mathcal{O}$ to generate actions, where $\mathcal{O}$ may include exteroceptive modalities (e.g., vision) that are complementary to the proprioceptive signals in $\mathbf{S}$.
\end{enumerate*}

This presents unique challenges for soft robots compared to rigid systems: the continuous deformation space $\mathbf{M}$ is theoretically infinite-dimensional as a continuum, demonstrations must account for the robot's inherent compliance, and the mapping between actuation and deformation is nonlinear or difficult to simulate in its entirety~\citep{liu2021influence}. The objective is to develop a framework that can effectively learn and execute manipulation skills while embracing these fundamental characteristics.




\section{Method}


\algname enables soft robot dexterous manipulation through a hierarchical approach that bridges the gap between kinesthetic teaching and autonomous execution. Our framework comprises five integrated components: a sensorized MOE-based soft hand with embedded strain sensors, a high-fidelity mesh-based shape estimation model, a shape-conditioned controller for precise trajectory tracking, a diffusion-based shape imitation policy, and a low-level controller that tracks desired shapes. By representing the soft hand's state through geometric mesh deformations rather than raw sensor values, \algname leverages the natural compliance of soft robots for dexterous skill learning while maintaining the precision necessary for complex manipulation tasks.

\subsection{MOE Soft Robot End-effector}
We leverage the multifinger omnidirectional end-effector (MOE) soft robot platform~\citept{yoo2025soft}, which comprises modular finger units that each operate independently. Two servo motors actuate each finger by applying tension to four tendons (Figure~\ref{figure:network}A). MOE has a modular design that allows the fingers to be rearranged into various configurations to suit specific task requirements. In experiments, we use a three-finger variant, inspired by research on object controllability using three-fingered rigid end-effectors~\citep{mason1985robot}.

Building on the original MOE finger design, we propose embedding low-cost conductive elastic rubber directly into the silicone elastomer body of each finger. The conductive rubber acts as sensors to measure deformation by varying their electrical resistance as they stretch, providing real-time proprioceptive feedback. Each finger incorporates four of these sensors, compactly positioned between the tendons.  A data acquisition (DAQ) circuit and board connected to the sensors can record resistance readings at approximately 400\,Hz. By seamlessly integrating the sensors into the elastomer during fabrication, we developed a fully deformable and sensorized finger body. The combined state spaces from all twelve (12) strain sensors provides our estimate $\mathbf{S}$ of the soft robot hand's true deformation state space $\mathbf{M}$ (Section \ref{sec:problem_statement}).


\subsection{Shape Estimation Model}
\label{sec:shape_estimation}
The shape estimation model maps strain sensor readings $\mathbf{R} \in \mathbb{R}^{n}$ to vertex displacements of the MOE fingers. Each of the three fingers contains four embedded strain sensors, giving $n = 12$ total resistance values. The model learns a function $f$ that predicts per-vertex deformations conditioned on both the sensor input and the undeformed mesh:
\[
f(\mathbf{R}, \{\mathbf{V}_{j,0}\}_{j=1}^3) = \{\Delta \mathbf{V}_j\}_{j=1}^3,
\]
where $\mathbf{V}_{j,0} \in \mathbb{R}^{N \times 3}$ denotes the initial vertex positions of finger $j$, and $\Delta \mathbf{V}_j$ are the predicted displacements. We implement $f$ using a FoldingNet-based architecture~\cite{yang2018foldingnet}. For each finger $j$, an encoder $h_\text{enc}$ transforms its four resistance values $\mathbf{R}_j$ into a latent code $\mathbf{z}_j \in \mathbb{R}^{128}$ as $\mathbf{z}_j = h_\text{enc}(\mathbf{R}_j)$.

A decoder $h_\text{dec}$ then predicts the displacement of each vertex $\mathbf{v}_{j,0}^i$ by combining its initial position with $\mathbf{z}_j$:
\[
\Delta \mathbf{v}_{j,i} = h_\text{dec}(\mathbf{v}_{j,0}^i, \mathbf{z}_j), \quad \mathbf{v}_{j,t}^i = \mathbf{v}_{j,0}^i + \Delta \mathbf{v}_{j,i}.
\]

This formulation allows the model to learn local deformation behavior conditioned on global strain input while preserving mesh topology. Predicting displacements rather than absolute positions improves stability and generalization as shown in comparison to baselines. The model is trained on a dataset of simulated deformations as described in the Appendix paired with corresponding simulated strain readings.

\subsection{Sim-to-Real Domain Alignment}
\label{sec: domain_align}
To bridge the gap between simulation and real-world deployment, we introduce a calibration procedure to align sensor readings from physical experiments with the simulated deformation model. Our goal is to determine a set of correction factors $\kappa_0, \dots, \kappa_{n-1}$, one per sensor, that align the strain-based deformation estimates in simulation with real-world resistance measurements.

We derive the following objective, based on the sensor model described in the Appendix, to align simulated deformations with real-world sensor readings: 
\begin{equation} \argmin_{\kappa_0, \dots, \kappa_{n-1}} \sum_{i=0}^{n-1} \sum_{t=0}^{T-1} \left( \sqrt{\frac{R_{i,t}}{R_{i,0}}} - \kappa_i \frac{L_{i,t}^S - L_{i,0}^S}{L_{i,0}^S} - 1 \right)^2, 
\end{equation} 
where $n$ is the number of sensors, $T$ is the number of sample frames, $R_{i,t}$ is the resistance measured from the $i$-th sensor at time step $t$, and $L_{i,t}^S$ is the corresponding simulated internal length of the $i$-th sensor at time $t$ The correction factors $\kappa_i$ are optimized to compensate for the domain mismatch between real and simulated sensor responses.



In practice, the true simulated lengths corresponding to each real-world resistance reading are unavailable, and the sensor-to-geometry mapping is not directly observable. Thus, we approximate this mapping by minimizing the unsupervised Chamfer distance between the observed 3D positions of points on the soft fingers and the predicted surface points from our shape estimation model:
\begin{equation}
\label{eq:chamfer_distance}
\begin{aligned}
    \mathcal{L}_{\text{UCD}} = \sum_{j=0}^{m-1} \sum_{\mathbf{p}_\text{obs} \in \mathcal{P}_\text{obs}^{(j)}} 
    \min_{\mathbf{p}_\text{pred} \in \mathcal{P}_\text{pred}^{(j)}} 
    \|\mathbf{p}_\text{obs} - \mathbf{p}_\text{pred}\|^2 ,
\end{aligned}
\end{equation}
\noindent where $\mathcal{P}_\text{obs}^{(j)}$ and $\mathcal{P}_\text{pred}^{(j)}$ denote the observed and predicted point clouds for the $j$-th example.

This alignment procedure ensures that the internal deformation estimated from resistance measurements produces a surface mesh consistent with external geometry observations. It connects to the overall pipeline by enabling the trained model to generalize from simulation to physical deployment without requiring paired real-world deformation labels.

\begin{figure*}[t]
    \centering
    \includegraphics[width=0.8\textwidth]{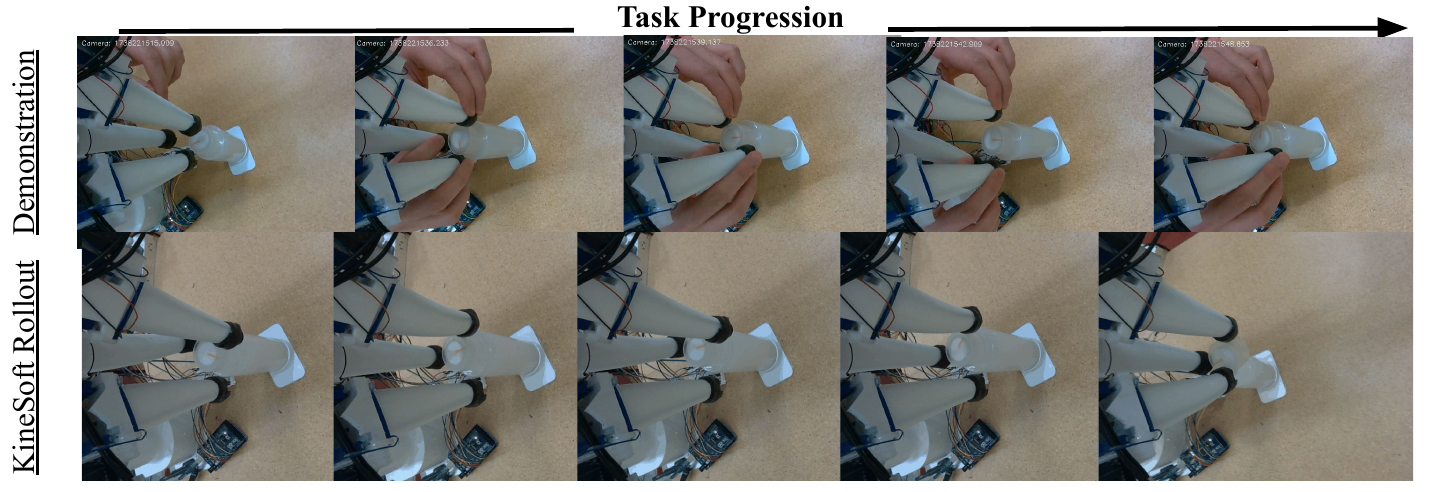} 
    \caption{\textbf{Demonstration and KineSoft Rollout for Bottle Unscrewing Task.}}
    \label{fig:rollout}
    \vspace{-0.5 cm}
\end{figure*}

\subsection{Shape-conditioned Controller}
\label{sec:shape_controller}

The shape-conditioned controller leverages real-time proprioceptive mesh state estimation of the MOE fingers to execute desired shape trajectories. For each finger $j$, the controller compares the current estimated vertex positions $\mathbf{V}_t$ with desired target positions $\mathbf{V}_t^D$ generated from the policy trajectory. 
%
%
Each finger is actuated by a pair of antagonistic tendons controlled by two servos. The actuation directions for each servo pair are represented by unit vectors $\mathbf{d}_{2j}, \mathbf{d}_{2j+1} \in \mathbb{R}^2$ that capture the primary deformation modes. The servo adjustments $\delta u_{j,t}$ for each finger are computed by projecting the shape error onto these actuation directions:
\[
\delta u_{j,t} = k_p \sum_{n} \mathbf{e}_{j,t}^n \cdot [\mathbf{d}_{2j}, \mathbf{d}_{2j+1}]^T,
\]
where $k_p$ is a scalar gain and $\mathbf{e}_{j,t}^n$ is the shape error for vertex $n$ defined by $\mathbf{e}_{j,t} = \mathbf{V}_{t,j}^D - \mathbf{V}_{t,j}$. In deployment,  the controller clips actions to prevent overloading the actuators. The controller runs at 100\,Hz with the shape estimation at each step, enabling responsive shape trajectory tracking. By projecting shape errors onto fitted actuation directions, the controller effectively translates desired deformations into appropriate servo commands despite the complex relationship between tendon actuation and finger deformation.

\subsection{Imitation Policy}


The shape estimation model (Section~\ref{sec:shape_estimation}) provides proprioceptive feedback via predicted surface vertex positions $\mathbf{V}_t$, while a wrist-mounted RGB-D camera captures external point cloud observations $\mathcal{P}_t$ of the workspace. We train a diffusion policy to imitate manipulation skills using these complementary inputs.

The policy predicts actions $a_t = \{\Delta \mathbf{V}_t, \Delta \mathbf{p}_t\}$ that couple surface deformations and end-effector motion. The state $s_t = \{h_\text{shape}(\mathbf{V}_t), h_\text{pc}(\mathcal{P}_t), \mathbf{p}_t\}$ combines proprioceptive and exteroceptive information: $h_\text{shape}$ is an MLP encoder applied to mesh vertices with temporal correspondence, $h_\text{pc}$ is a DP3 encoder for the RGB-D point cloud, and $\mathbf{p}_t$ is the current end-effector pose.

The diffusion-based imitation policy learns to denoise a vertex-level action trajectory via the reverse process $a_{t-1} = \mu_\theta(a_t, s_t, t) + \sigma_t \mathbf{z}$, where $\mu_\theta$ is a learned denoising network and $\mathbf{z} \sim \mathcal{N}(0, \mathbf{I})$. Combined with the shape-conditioned controller (Section~\ref{sec:shape_controller}), the policy enables robust, contact-rich manipulation by fusing learned high-level intent with precise low-level tracking.


\section{Experiments}
\label{sec:experiments}
We evaluate \algname through real-world experiments that assess each component of the framework and key design choices. We quantify the accuracy of the shape estimation model, followed by evaluating the shape-conditioned controller's ability to track target deformations. Finally, we assess \algname policies across six in-hand manipulation tasks involving both rigid and deformable objects. These experiments demonstrate the effectiveness of shape-based representations for proprioceptive control and reliable manipulation, particularly in settings where soft robots offer inherent advantages.



\vspace{-0.5cm}
\subsection{Shape Estimation}
\begin{wraptable}{r}{0.45\textwidth}
    \centering
    \caption{Shape Estimation Fidelity}
    \resizebox{0.9\linewidth}{!}{%
    \begin{tabular}{lS[table-format=1.2(1.2)]}
        \toprule
        Method & {Shape Error [mm]} \\
        \midrule
        PneuFlex Sensor & 3.70 (1.36)  \\
        DeepSoRo & 3.27 (1.05)  \\
        \textbf{KineSoft} (naive) & 4.91 (2.85)\\
        \textbf{KineSoft} (unconstrained) & 4.36 (3.47)\\
        \textbf{KineSoft} (model) & \bfseries 1.92 (0.39) \\
        \bottomrule
    \end{tabular}
    }
    \label{tab:shape_results}
\end{wraptable}
We evaluated the proposed shape estimation model against baselines from the literature: the constant curvature model~\citep{della2020improved, yoo2021analytical}, an analytical representation of soft robot deformation, and DeepSoRo~\citep{wang2020real}, a learning-based point cloud reconstruction method. In addition, we compared against two ablations of our approach: a naively calibrated linear mapping and an unconstrained fully connected model, both trained using the domain alignment procedure described in Section~\ref{sec: domain_align}. Evaluation was conducted using observed point clouds captured from the physical setup shown in Figure~\ref{figure:calibration_setup}, and performance was measured using the unidirectional Chamfer distance defined in Equation~\ref{eq:chamfer_distance}. Quantitative results are summarized in Table~\ref{tab:shape_results} and visualized in Figure~\ref{figure:trajectory_tracking_results}. Our model achieved a shape estimation error of 1.92,mm, representing a 41.3\% improvement over the best baseline (DeepSoRo) and a 60.9\% improvement over the linear variant of our method.


\vspace{-0.3cm}
\subsection{Shape-conditioned Controller Performance}

\begin{wraptable}{r}{0.45\textwidth}
    \caption{Shape Tracking Error Comparison}
    \label{tab:shape_tracking}
    \resizebox{\linewidth}{!}{
    \begin{tabular}{lcS[table-format=1.2(1.2)]}
        \toprule
        Method & Representation & {Error [mm]} \\
        \midrule
        Strain-tracking & Strain & 6.20 (2.39) \\
        \textbf{KineSoft} & Mesh & \bfseries 3.29 (0.91)\\
        \bottomrule
    \end{tabular}
    }
\end{wraptable}

We evaluated the proposed shape-conditioned controller against a strain-tracking baseline that directly uses sensor readings for control as implemented in prior soft robot manipulation works~\citep{bhatt2022surprisingly,sieler2023dexterous}. For evaluation, we collected reference trajectories through kinesthetic teaching, where a demonstrator physically deformed the fingers into desired configurations. During execution, the controllers had to track these trajectories using tendon actuation. This evaluation highlights a fundamental challenge in soft robot imitation: the sensor signals generated during kinesthetic demonstration (when fingers are manually deformed) differ significantly from those produced during autonomous execution (when tendons are actuated). The strain-tracking baseline, which attempts to directly match these sensor readings, struggles with this demonstration-execution gap, achieving 6.20\,mm tracking error. In contrast, the proposed shape-conditioned controller bridges this gap by tracking the geometric shape itself rather than the underlying sensor signals, achieving 3.29\,mm error, a 47\,\% improvement.

\begin{figure}[t]
\centering
\includegraphics[width=0.9\columnwidth]{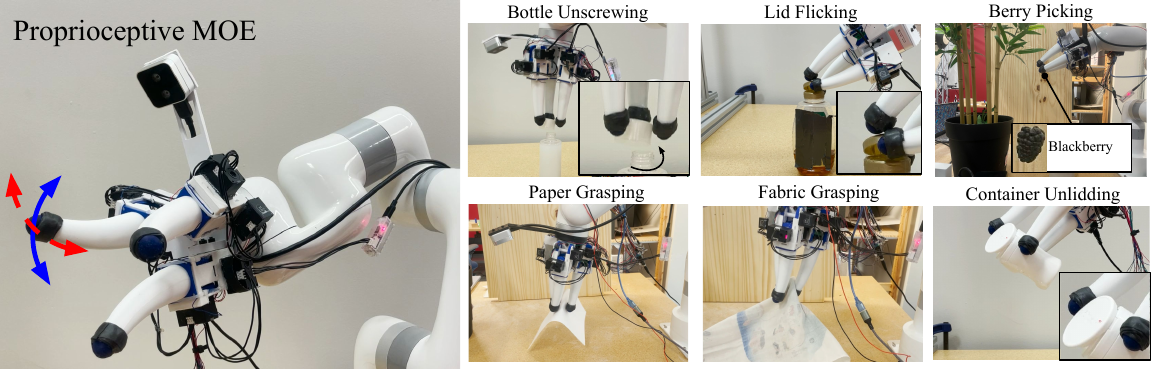}
\caption{\textbf{Tasks.} We evaluate the performance of the shape-based \algname policy across six manipulation tasks. These tasks highlight the advantages of soft robotic hands in contact-rich and delicate object interaction. Red arrows indicate actuation directions for Servo 1; blue arrows indicate actuation directions for Servo 2.}
\vspace{-0.6 cm}
\label{figure:task_results}
\end{figure}



\subsection{In-hand Manipulation Task Performance}
We evaluated KineSoft on six in-hand manipulation tasks (Figure~\ref{figure:task_results}) to showcase both the strengths of soft robots, such as delicate object handling~\cite{sinatra_ultragentle_2019,bauer2023soft,a_l_gunderman_tendon-driven_2022} and contact-rich interaction~\cite{yoo2025soft, pall2021analysis}, and their conventional challenges, including tasks requiring accurate multi-finger coordination~\cite{hughes2016soft}.
\begin{description}[wide, labelwidth=!, labelindent=0pt]
    \item[Bottle Unscrewing:] MOE must rotate and lift a bottle cap using coordinated finger motion. 
    \item[Lid Flicking:] MOE flicks open a hinged lid on a container.
    \item[Berry Picking:] MOE gently picks a berry from a soft branch, evaluating precise grasping and compliant contact with fragile objects inspired by prior works on soft robot fruit harvesting~\cite{a_l_gunderman_tendon-driven_2022}.
    \item[Paper Grasping:] MOE initiates contact with and lifts a flat sheet of paper from a table.
    \item[Fabric Grasping:] MOE conforms around and lifts a loosely draped piece of fabric, highlighting contact-rich manipulation of deformable objects.
    \item[Container Unlidding:] MOE grasps and opens the snap-top lid of a plastic container while maintaining a stable grasp. The task highlights multi-finger coordination and localized force application.
\end{description}
Each task was performed in 20 real-world trials using demonstrations collected via kinesthetic teaching, where the operator directly deforms the soft fingers. During execution, KineSoft reproduces these deformations through tendon actuation, bridging the sim-to-real gap with learned shape-aware control. Table~\ref{tab:task_performance} reports success rates across all six tasks. KineSoft achieved an 85\,\% success rate on Bottle Unscrewing, compared to 0\,\% for a strain-matching baseline based on a state-of-the-art soft in-hand manipulation method~\cite{sieler2023dexterous}, implemented with a diffusion policy and trained on the same demonstration data. On Lid Flicking, KineSoft reached 100\,\% success, while the baseline achieved 90\,\%. For contact-sensitive tasks such as Berry Picking, Fabric Pickup, and Container Unlidding, KineSoft achieved success rates between 70\,\% and 80\,\%, whereas the baseline remained below 35\,\% on all three. These results highlight the effectiveness of shape-based representations for robust control and reinforce the advantages of soft hands in contact-rich and delicate manipulation settings.



\begin{table}[t]
\centering
\caption{Task success rates on 6 different tasks over 20 trials each comparing the baseline strain-matching method to the proposed KineSoft.}
\label{tab:task_performance}
\begin{tabular}{lcccccc}
\toprule
 & Bottle & Lid & Berry & Paper & Fabric & Container\\
\bf Method & Unscrewing & Flicking  & Picking & Grasping & Pickup & Unlidding \\
\midrule
Strain Policy & \phantom00/20 & 18/20 & \phantom07/20 & 13/20 & \phantom 01/20 & \phantom03/20  \\ 
KineSoft & \bf 17/20 & \bf 20/20 & \bf 16/20 & \bf 19/20 & \bf 13/20 & \bf 14/20 \\
\bottomrule \\[-6pt]
\end{tabular}
\vspace{-0.5cm}
\end{table}

%

\vspace{-0.2cm}
\section{Conclusion and Lessons}
\vspace{-0.2cm}
This paper presents KineSoft, a framework for learning dexterous manipulation skills with soft robot hands that embraces, rather than fights against, their inherent compliance. The key insight is recognizing that while this compliance enables intuitive kinesthetic teaching, it creates a fundamental gap between demonstration and execution, where the deformations and sensor signals during human demonstration differ substantially from those during autonomous execution through tendon actuation.KineSoft addresses this challenge through a hierarchical approach. A shape estimation model provides consistent geometric representations across demonstration and execution modes while a domain alignment method enables robust transfer of simulation-trained models to real hardware and a shape-conditioned controller reliably tracks the policy's generated deformation trajectories despite the different underlying actuation mechanisms. Then, a high-level imitation policy learns to generate target vertex deformations from demonstrations, capturing the intended manipulation strategy in geometric grounding. In experiments, results demonstrate that this shape-based hierarchical approach enables more effective skill transfer than methods that attempt to directly match sensor signals or joint configurations. This work suggests that successful imitation learning for dexterous soft robots requires careful consideration of how demonstration and execution modes differ, and appropriate intermediate representations to bridge this gap.

\section{Limitations}

While KineSoft demonstrates effective shape estimation and trajectory tracking for soft robotic manipulation, several limitations remain. First, the current system lacks explicit force-feedback mechanisms, relying instead on kinematic trajectories and deformation states to encode manipulation skills. This limits precise force control during object interactions. In contrast, human manipulation integrates both proprioception and tactile sensing, suggesting potential benefits from adding complementary tactile input. 

Additionally, although kinesthetic teaching provides an robust approach to demonstrating some dexterous tasks, it requires the expert demonstrator to recognize the MOE finger's workspace limitations to avoid demonstrating shapes that the MOE cannot reach. In future work, we aim to leverage KineSoft's real-time shape estimation to communicate robot affordances to users via a visual user interface during demonstrations.

Although the sensors used in this work are readily available and low-cost, they introduce fabrication complexity when integrating into the prior MOE design, which was relatively simple to fabricate. We also observed that the sensors may not respond dynamically in fast manipulation tasks due to commonly observed soft body phenomena such as creep and hysteresis, where the sensor signals require some time to reach equilibrium conditions when stretched quickly. For durability, we included results on long-term continuous operation and sensor signal deviation in the Appendix.

KineSoft's shape-conditioned low-level controller weighs the errors from the vertices of the reference meshes equally. Implicitly, the controller prioritizes the vertices closer to the tip higher because the displacement tends to be greater further away from the base. For many tasks, this is an appropriate weighing as the finger tips play a vital role in performing dexterous in-hand manipulation tasks~\citep{bullock2014analyzing}. However, some tasks may require different regions of the fingers to be prioritized based on contact probability. 

\bibliography{references}  

\begin{thebibliography}{60}
\providecommand{\natexlab}[1]{#1}
\providecommand{\url}[1]{\texttt{#1}}
\expandafter\ifx\csname urlstyle\endcsname\relax
  \providecommand{\doi}[1]{doi: #1}\else
  \providecommand{\doi}{doi: \begingroup \urlstyle{rm}\Url}\fi

\bibitem[Yoo et~al.(2025)Yoo, Dennler, Xing, Matari{\'c}, Nikolaidis, Ichnowski, and Oh]{yoo2025soft}
U.~Yoo, N.~Dennler, E.~Xing, M.~Matari{\'c}, S.~Nikolaidis, J.~Ichnowski, and J.~Oh.
\newblock Soft and compliant contact-rich hair manipulation and care.
\newblock In \emph{Proc. of IEEE International Conference on Human-Robot Interaction (HRI)}, 2025.

\bibitem[Gilday et~al.(2024)Gilday, Zubak, Raabe, and Hughes]{gilday2024rigid}
K.~Gilday, I.~Zubak, A.~Raabe, and J.~Hughes.
\newblock From rigid to soft robotic approaches for minimally invasive neurosurgery.
\newblock \emph{arXiv preprint arXiv:2404.14071}, 2024.

\bibitem[Bhatt et~al.(2022)Bhatt, Sieler, Puhlmann, and Brock]{bhatt2022surprisingly}
A.~Bhatt, A.~Sieler, S.~Puhlmann, and O.~Brock.
\newblock Surprisingly robust in-hand manipulation: An empirical study.
\newblock \emph{arXiv preprint arXiv:2201.11503}, 2022.

\bibitem[Homberg et~al.(2019)Homberg, Katzschmann, Dogar, and Rus]{homberg2019robust}
B.~S. Homberg, R.~K. Katzschmann, M.~R. Dogar, and D.~Rus.
\newblock Robust proprioceptive grasping with a soft robot hand.
\newblock \emph{Autonomous robots}, 43:\penalty0 681--696, 2019.

\bibitem[Yao et~al.(2024)Yao, Yoo, Oh, Atkeson, and Ichnowski]{yao2024soft}
Y.~Yao, U.~Yoo, J.~Oh, C.~G. Atkeson, and J.~Ichnowski.
\newblock Soft robotic dynamic in-hand pen spinning.
\newblock \emph{arXiv preprint arXiv:2411.12734}, 2024.

\bibitem[Firth et~al.(2022)Firth, Dunn, Haeusler, and Sun]{firth2022anthropomorphic}
C.~Firth, K.~Dunn, M.~H. Haeusler, and Y.~Sun.
\newblock Anthropomorphic soft robotic end-effector for use with collaborative robots in the construction industry.
\newblock \emph{Automation in Construction}, 138:\penalty0 104218, 2022.

\bibitem[Abondance et~al.(2020)Abondance, Teeple, and Wood]{abondance2020dexterous}
S.~Abondance, C.~B. Teeple, and R.~J. Wood.
\newblock A dexterous soft robotic hand for delicate in-hand manipulation.
\newblock \emph{IEEE Robotics and Automation Letters}, 5\penalty0 (4):\penalty0 5502--5509, 2020.

\bibitem[Jitosho et~al.(2025)Jitosho, Winston, Yang, Li, Ahlquist, Woehrle, Liu, and Okamura]{jitosho2025flying}
R.~Jitosho, C.~E. Winston, S.~Yang, J.~Li, M.~Ahlquist, N.~J. Woehrle, C.~K. Liu, and A.~M. Okamura.
\newblock Flying vines: Design, modeling, and control of a soft aerial robotic arm.
\newblock \emph{arXiv preprint arXiv:2503.20754}, 2025.

\bibitem[Chi et~al.(2023)Chi, Xu, Feng, Cousineau, Du, Burchfiel, Tedrake, and Song]{chi2023diffusion}
C.~Chi, Z.~Xu, S.~Feng, E.~Cousineau, Y.~Du, B.~Burchfiel, R.~Tedrake, and S.~Song.
\newblock Diffusion policy: Visuomotor policy learning via action diffusion.
\newblock \emph{The International Journal of Robotics Research}, page 02783649241273668, 2023.

\bibitem[Memmel et~al.(2025)Memmel, Berg, Chen, Gupta, and Francis]{memmel2025strap}
M.~Memmel, J.~Berg, B.~Chen, A.~Gupta, and J.~Francis.
\newblock {STRAP}: Robot sub-trajectory retrieval for augmented policy learning.
\newblock In \emph{The Thirteenth International Conference on Learning Representations}, 2025.
\newblock URL \url{https://openreview.net/pdf?id=4VHiptx7xe}.

\bibitem[Hu et~al.(2023)Hu, Xie, Jain, Francis, Patrikar, Keetha, Kim, Xie, Zhang, Fang, et~al.]{hu2023toward}
Y.~Hu, Q.~Xie, V.~Jain, J.~Francis, J.~Patrikar, N.~Keetha, S.~Kim, Y.~Xie, T.~Zhang, H.-S. Fang, et~al.
\newblock Toward general-purpose robots via foundation models: A survey and meta-analysis.
\newblock \emph{arXiv preprint arXiv:2312.08782}, 2023.

\bibitem[Fu et~al.(2024)Fu, Zhao, and Finn]{fu2024mobile}
Z.~Fu, T.~Z. Zhao, and C.~Finn.
\newblock Mobile aloha: Learning bimanual mobile manipulation with low-cost whole-body teleoperation.
\newblock \emph{arXiv preprint arXiv:2401.02117}, 2024.

\bibitem[Yoo et~al.(2024)Yoo, Hung, Francis, Oh, and Ichnowski]{yoo2024ropotter}
U.~Yoo, A.~Hung, J.~Francis, J.~Oh, and J.~Ichnowski.
\newblock Ropotter: Toward robotic pottery and deformable object manipulation with structural priors.
\newblock In \emph{2024 IEEE-RAS 23rd International Conference on Humanoid Robots (Humanoids)}, pages 843--850. IEEE, 2024.

\bibitem[Qi et~al.(2025)Qi, Yi, Lambeta, Ma, Calandra, and Malik]{qi2025simple}
H.~Qi, B.~Yi, M.~Lambeta, Y.~Ma, R.~Calandra, and J.~Malik.
\newblock From simple to complex skills: The case of in-hand object reorientation.
\newblock \emph{arXiv preprint arXiv:2501.05439}, 2025.

\bibitem[Wang et~al.(2024)Wang, Yuan, Che, Qi, Ma, Malik, and Wang]{wang2024penspin}
J.~Wang, Y.~Yuan, H.~Che, H.~Qi, Y.~Ma, J.~Malik, and X.~Wang.
\newblock Lessons from learning to spin “pens”.
\newblock In \emph{CoRL}, 2024.

\bibitem[Qin et~al.(2023)Qin, Yang, Huang, Van~Wyk, Su, Wang, Chao, and Fox]{qin2023anyteleop}
Y.~Qin, W.~Yang, B.~Huang, K.~Van~Wyk, H.~Su, X.~Wang, Y.-W. Chao, and D.~Fox.
\newblock Anyteleop: A general vision-based dexterous robot arm-hand teleoperation system.
\newblock \emph{arXiv preprint arXiv:2307.04577}, 2023.

\bibitem[Ze et~al.(2024)Ze, Zhang, Zhang, Hu, Wang, and Xu]{ze20243d}
Y.~Ze, G.~Zhang, K.~Zhang, C.~Hu, M.~Wang, and H.~Xu.
\newblock 3d diffusion policy.
\newblock \emph{arXiv preprint arXiv:2403.03954}, 2024.

\bibitem[Wang et~al.(2023)Wang, Lam, Chen, Li, Zhang, Su, and Wang]{wang2023soft}
L.~Wang, J.~Lam, X.~Chen, J.~Li, R.~Zhang, Y.~Su, and Z.~Wang.
\newblock Soft robot proprioception using unified soft body encoding and recurrent neural network.
\newblock \emph{Soft Robotics}, 10\penalty0 (4):\penalty0 825--837, 2023.

\bibitem[Yoo et~al.(2023)Yoo, Zhao, Altamirano, Yuan, and Feng]{yoo2023toward}
U.~Yoo, H.~Zhao, A.~Altamirano, W.~Yuan, and C.~Feng.
\newblock Toward zero-shot sim-to-real transfer learning for pneumatic soft robot 3d proprioceptive sensing.
\newblock In \emph{2023 IEEE International Conference on Robotics and Automation (ICRA)}, pages 544--551. IEEE, 2023.

\bibitem[Yoo et~al.(2024)Yoo, Lopez, Ichnowski, and Oh]{yoo2024poe}
U.~Yoo, Z.~Lopez, J.~Ichnowski, and J.~Oh.
\newblock Poe: Acoustic soft robotic proprioception for omnidirectional end-effectors.
\newblock \emph{arXiv preprint arXiv:2401.09382}, 2024.

\bibitem[Andrychowicz et~al.(2020)Andrychowicz, Baker, Chociej, Jozefowicz, McGrew, Pachocki, Petron, Plappert, Powell, Ray, et~al.]{andrychowicz2020learning}
O.~M. Andrychowicz, B.~Baker, M.~Chociej, R.~Jozefowicz, B.~McGrew, J.~Pachocki, A.~Petron, M.~Plappert, G.~Powell, A.~Ray, et~al.
\newblock Learning dexterous in-hand manipulation.
\newblock \emph{The International Journal of Robotics Research}, 39\penalty0 (1):\penalty0 3--20, 2020.

\bibitem[Della~Santina et~al.(2023)Della~Santina, Duriez, and Rus]{della2023model}
C.~Della~Santina, C.~Duriez, and D.~Rus.
\newblock Model-based control of soft robots: A survey of the state of the art and open challenges.
\newblock \emph{IEEE Control Systems Magazine}, 43\penalty0 (3):\penalty0 30--65, 2023.

\bibitem[Thuruthel et~al.(2018)Thuruthel, Falotico, Renda, and Laschi]{thuruthel2018model}
T.~G. Thuruthel, E.~Falotico, F.~Renda, and C.~Laschi.
\newblock Model-based reinforcement learning for closed-loop dynamic control of soft robotic manipulators.
\newblock \emph{IEEE Transactions on Robotics}, 35\penalty0 (1):\penalty0 124--134, 2018.

\bibitem[Schegg et~al.(2023)Schegg, M{\'e}nager, Khairallah, Marchal, Dequidt, Preux, and Duriez]{schegg2023sofagym}
P.~Schegg, E.~M{\'e}nager, E.~Khairallah, D.~Marchal, J.~Dequidt, P.~Preux, and C.~Duriez.
\newblock Sofagym: An open platform for reinforcement learning based on soft robot simulations.
\newblock \emph{Soft Robotics}, 10\penalty0 (2):\penalty0 410--430, 2023.

\bibitem[Bhagat et~al.(2019)Bhagat, Banerjee, Ho~Tse, and Ren]{bhagat2019deep}
S.~Bhagat, H.~Banerjee, Z.~T. Ho~Tse, and H.~Ren.
\newblock Deep reinforcement learning for soft, flexible robots: Brief review with impending challenges.
\newblock \emph{Robotics}, 8\penalty0 (1):\penalty0 4, 2019.

\bibitem[Johns(2021)]{johns2021coarse}
E.~Johns.
\newblock Coarse-to-fine imitation learning: Robot manipulation from a single demonstration.
\newblock In \emph{2021 IEEE international conference on robotics and automation (ICRA)}, pages 4613--4619. IEEE, 2021.

\bibitem[Arunachalam et~al.(2023{\natexlab{a}})Arunachalam, G{\"u}zey, Chintala, and Pinto]{arunachalam2023holo}
S.~P. Arunachalam, I.~G{\"u}zey, S.~Chintala, and L.~Pinto.
\newblock Holo-dex: Teaching dexterity with immersive mixed reality.
\newblock In \emph{2023 IEEE International Conference on Robotics and Automation (ICRA)}, pages 5962--5969. IEEE, 2023{\natexlab{a}}.

\bibitem[Arunachalam et~al.(2023{\natexlab{b}})Arunachalam, Silwal, Evans, and Pinto]{arunachalam2023dexterous}
S.~P. Arunachalam, S.~Silwal, B.~Evans, and L.~Pinto.
\newblock Dexterous imitation made easy: A learning-based framework for efficient dexterous manipulation.
\newblock In \emph{2023 ieee international conference on robotics and automation (icra)}, pages 5954--5961. IEEE, 2023{\natexlab{b}}.

\bibitem[Wei and Xu(2024)]{wei2024wearable}
D.~Wei and H.~Xu.
\newblock A wearable robotic hand for hand-over-hand imitation learning.
\newblock In \emph{2024 IEEE International Conference on Robotics and Automation (ICRA)}, pages 18113--18119. IEEE, 2024.

\bibitem[Weinberg et~al.(2024)Weinberg, Shirizly, Azulay, and Sintov]{weinberg2024survey}
A.~I. Weinberg, A.~Shirizly, O.~Azulay, and A.~Sintov.
\newblock Survey of learning-based approaches for robotic in-hand manipulation.
\newblock \emph{Frontiers in Robotics and AI}, 11:\penalty0 1455431, 2024.

\bibitem[Rus and Tolley(2015)]{rus2015design}
D.~Rus and M.~T. Tolley.
\newblock Design, fabrication and control of soft robots.
\newblock \emph{Nature}, 521\penalty0 (7553):\penalty0 467--475, 2015.

\bibitem[Zhou et~al.(2023)Zhou, Chen, and Yang]{zhou2023soft}
X.~Zhou, X.~Chen, and T.~Yang.
\newblock Soft robotic grippers.
\newblock \emph{Advanced Intelligent Systems}, 5\penalty0 (1):\penalty0 2000198, 2023.

\bibitem[Liu et~al.(2024)Liu, Puthuveetil, Padmanabha, Khokar, Temel, and Erickson]{liu2024skingrip}
F.~Liu, K.~Puthuveetil, A.~Padmanabha, K.~Khokar, Z.~Temel, and Z.~Erickson.
\newblock Skingrip: An adaptive soft robotic manipulator with capacitive sensing for whole-limb bathing assistance.
\newblock \emph{arXiv preprint arXiv:2405.02772}, 2024.

\bibitem[Puhlmann et~al.(2022)Puhlmann, Harris, and Brock]{puhlmann2022rbo}
S.~Puhlmann, J.~Harris, and O.~Brock.
\newblock Rbo hand 3: A platform for soft dexterous manipulation.
\newblock \emph{IEEE Transactions on Robotics}, 38\penalty0 (6):\penalty0 3434--3449, 2022.

\bibitem[Yasa et~al.(2023)Yasa, Toshimitsu, Michelis, Jones, Filippi, Buchner, and Katzschmann]{yasa2023overview}
O.~Yasa, Y.~Toshimitsu, M.~Y. Michelis, L.~S. Jones, M.~Filippi, T.~Buchner, and R.~K. Katzschmann.
\newblock An overview of soft robotics.
\newblock \emph{Annual Review of Control, Robotics, and Autonomous Systems}, 6\penalty0 (1):\penalty0 1--29, 2023.

\bibitem[Gupta et~al.(2016)Gupta, Eppner, Levine, and Abbeel]{gupta2016learning}
A.~Gupta, C.~Eppner, S.~Levine, and P.~Abbeel.
\newblock Learning dexterous manipulation for a soft robotic hand from human demonstrations.
\newblock In \emph{2016 IEEE/RSJ International Conference on Intelligent Robots and Systems (IROS)}, pages 3786--3793. IEEE, 2016.

\bibitem[Pagoli et~al.(2021)Pagoli, Chapelle, Corrales, Mezouar, and Lapusta]{pagoli2021soft}
A.~Pagoli, F.~Chapelle, J.~A. Corrales, Y.~Mezouar, and Y.~Lapusta.
\newblock A soft robotic gripper with an active palm and reconfigurable fingers for fully dexterous in-hand manipulation.
\newblock \emph{IEEE Robotics and Automation Letters}, 6\penalty0 (4):\penalty0 7706--7713, 2021.

\bibitem[Zhou et~al.(2024)Zhou, Li, Wang, and Lyu]{zhou2024integrated}
S.~Zhou, Y.~Li, Q.~Wang, and Z.~Lyu.
\newblock Integrated actuation and sensing: Toward intelligent soft robots.
\newblock \emph{Cyborg and Bionic Systems}, 5:\penalty0 0105, 2024.

\bibitem[Ku et~al.(2024)Ku, Song, Park, Lee, and Park]{ku2024soft}
S.~Ku, B.-H. Song, T.~Park, Y.~Lee, and Y.-L. Park.
\newblock Soft modularized robotic arm for safe human--robot interaction based on visual and proprioceptive feedback.
\newblock \emph{The International Journal of Robotics Research}, 43\penalty0 (8):\penalty0 1128--1150, 2024.

\bibitem[Stella et~al.(2023)Stella, Guan, Della~Santina, and Hughes]{stella2023piecewise}
F.~Stella, Q.~Guan, C.~Della~Santina, and J.~Hughes.
\newblock Piecewise affine curvature model: a reduced-order model for soft robot-environment interaction beyond pcc.
\newblock In \emph{2023 IEEE International Conference on Soft Robotics (RoboSoft)}, pages 1--7. IEEE, 2023.

\bibitem[Yoo et~al.(2021)Yoo, Liu, Deshpande, and Alamabeigi]{yoo2021analytical}
U.~Yoo, Y.~Liu, A.~D. Deshpande, and F.~Alamabeigi.
\newblock Analytical design of a pneumatic elastomer robot with deterministically adjusted stiffness.
\newblock \emph{IEEE robotics and automation letters}, 6\penalty0 (4):\penalty0 7773--7780, 2021.

\bibitem[Wall et~al.(2017)Wall, Z{\"o}ller, and Brock]{wall2017method}
V.~Wall, G.~Z{\"o}ller, and O.~Brock.
\newblock A method for sensorizing soft actuators and its application to the rbo hand 2.
\newblock In \emph{2017 IEEE International Conference on Robotics and Automation (ICRA)}, pages 4965--4970. IEEE, 2017.

\bibitem[Liu et~al.(2021)Liu, Yoo, Ha, Atashzar, and Alambeigi]{liu2021influence}
Y.~Liu, U.~Yoo, S.~Ha, S.~F. Atashzar, and F.~Alambeigi.
\newblock Influence of antagonistic tensions on distributed friction forces of multisegment tendon-driven continuum manipulators with irregular geometry.
\newblock \emph{IEEE/ASME Transactions on Mechatronics}, 27\penalty0 (5):\penalty0 2418--2428, 2021.

\bibitem[Liu et~al.(2020)Liu, Ahn, Yoo, Cohen, and Alambeigi]{liu2020toward}
Y.~Liu, S.~H. Ahn, U.~Yoo, A.~R. Cohen, and F.~Alambeigi.
\newblock Toward analytical modeling and evaluation of curvature-dependent distributed friction force in tendon-driven continuum manipulators.
\newblock In \emph{2020 IEEE/RSJ International Conference on Intelligent Robots and Systems (IROS)}, pages 8823--8828. IEEE, 2020.

\bibitem[Tapia et~al.(2020)Tapia, Knoop, Mutn{\`y}, Otaduy, and B{\"a}cher]{tapia2020makesense}
J.~Tapia, E.~Knoop, M.~Mutn{\`y}, M.~A. Otaduy, and M.~B{\"a}cher.
\newblock Makesense: Automated sensor design for proprioceptive soft robots.
\newblock \emph{Soft robotics}, 7\penalty0 (3):\penalty0 332--345, 2020.

\bibitem[Katzschmann et~al.(2019)Katzschmann, Della~Santina, Toshimitsu, Bicchi, and Rus]{katzschmann2019dynamic}
R.~K. Katzschmann, C.~Della~Santina, Y.~Toshimitsu, A.~Bicchi, and D.~Rus.
\newblock Dynamic motion control of multi-segment soft robots using piecewise constant curvature matched with an augmented rigid body model.
\newblock In \emph{2019 2nd IEEE International Conference on Soft Robotics (RoboSoft)}, pages 454--461. IEEE, 2019.

\bibitem[Mason and Salisbury~Jr(1985)]{mason1985robot}
M.~T. Mason and J.~K. Salisbury~Jr.
\newblock Robot hands and the mechanics of manipulation.
\newblock \emph{The MIT Press, Cambridge, MA}, 1985.

\bibitem[Yang et~al.(2018)Yang, Feng, Shen, and Tian]{yang2018foldingnet}
Y.~Yang, C.~Feng, Y.~Shen, and D.~Tian.
\newblock Foldingnet: Point cloud auto-encoder via deep grid deformation.
\newblock In \emph{Proceedings of the IEEE conference on computer vision and pattern recognition}, pages 206--215, 2018.

\bibitem[Della~Santina et~al.(2020)Della~Santina, Bicchi, and Rus]{della2020improved}
C.~Della~Santina, A.~Bicchi, and D.~Rus.
\newblock On an improved state parametrization for soft robots with piecewise constant curvature and its use in model based control.
\newblock \emph{IEEE Robotics and Automation Letters}, 5\penalty0 (2):\penalty0 1001--1008, 2020.

\bibitem[Wang et~al.(2020)Wang, Wang, Du, Xiao, Yuan, and Feng]{wang2020real}
R.~Wang, S.~Wang, S.~Du, E.~Xiao, W.~Yuan, and C.~Feng.
\newblock Real-time soft body 3d proprioception via deep vision-based sensing.
\newblock \emph{IEEE Robotics and Automation Letters}, 5\penalty0 (2):\penalty0 3382--3389, 2020.

\bibitem[Sieler and Brock(2023)]{sieler2023dexterous}
A.~Sieler and O.~Brock.
\newblock Dexterous soft hands linearize feedback-control for in-hand manipulation.
\newblock In \emph{2023 IEEE/RSJ International Conference on Intelligent Robots and Systems (IROS)}, pages 8757--8764. IEEE, 2023.

\bibitem[Sinatra et~al.(2019)Sinatra, Teeple, Vogt, Parker, Gruber, and Wood]{sinatra_ultragentle_2019}
N.~R. Sinatra, C.~B. Teeple, D.~M. Vogt, K.~K. Parker, D.~F. Gruber, and R.~J. Wood.
\newblock Ultragentle manipulation of delicate structures using a soft robotic gripper.
\newblock \emph{Science Robotics}, 4\penalty0 (33):\penalty0 eaax5425, Aug. 2019.
\newblock \doi{10.1126/scirobotics.aax5425}.
\newblock URL \url{https://doi.org/10.1126/scirobotics.aax5425}.
\newblock Publisher: American Association for the Advancement of Science.

\bibitem[Bauer et~al.(2023)Bauer, Bauer, and Pollard]{bauer2023soft}
D.~Bauer, C.~Bauer, and N.~S. Pollard.
\newblock Soft robotic end-effectors in the wild: A case study of a soft manipulator for green bell pepper harvesting.
\newblock In \emph{2nd AAAI Workshop on AI for Agriculture and Food Systems}, 2023.

\bibitem[{A. L. Gunderman} et~al.(2022){A. L. Gunderman}, {J. A. Collins}, {A. L. Myers}, {R. T. Threlfall}, and {Y. Chen}]{a_l_gunderman_tendon-driven_2022}
{A. L. Gunderman}, {J. A. Collins}, {A. L. Myers}, {R. T. Threlfall}, and {Y. Chen}.
\newblock Tendon-{Driven} {Soft} {Robotic} {Gripper} for {Blackberry} {Harvesting}.
\newblock \emph{IEEE Robotics and Automation Letters}, 7\penalty0 (2):\penalty0 2652--2659, Apr. 2022.
\newblock ISSN 2377-3766.
\newblock \doi{10.1109/LRA.2022.3143891}.

\bibitem[P{\'a}ll and Brock(2021)]{pall2021analysis}
E.~P{\'a}ll and O.~Brock.
\newblock Analysis of open-loop grasping from piles.
\newblock In \emph{2021 IEEE International Conference on Robotics and Automation (ICRA)}, pages 2591--2597. IEEE, 2021.

\bibitem[Hughes et~al.(2016)Hughes, Culha, Giardina, Guenther, Rosendo, and Iida]{hughes2016soft}
J.~Hughes, U.~Culha, F.~Giardina, F.~Guenther, A.~Rosendo, and F.~Iida.
\newblock Soft manipulators and grippers: A review.
\newblock \emph{Frontiers in Robotics and AI}, 3:\penalty0 69, 2016.

\bibitem[Bullock et~al.(2014)Bullock, Feix, and Dollar]{bullock2014analyzing}
I.~M. Bullock, T.~Feix, and A.~M. Dollar.
\newblock Analyzing human fingertip usage in dexterous precision manipulation: Implications for robotic finger design.
\newblock In \emph{2014 IEEE/RSJ International Conference on Intelligent Robots and Systems}, pages 1622--1628. IEEE, 2014.

\bibitem[Starkova and Aniskevich(2010)]{starkova2010poisson}
O.~Starkova and A.~Aniskevich.
\newblock Poisson's ratio and the incompressibility relation for various strain measures with the example of a silica-filled sbr rubber in uniaxial tension tests.
\newblock \emph{Polymer Testing}, 29\penalty0 (3):\penalty0 310--318, 2010.

\bibitem[Zhang et~al.(2023)Zhang, Wang, Truby, Chin, and Rus]{zhang2023machine}
A.~Zhang, T.-H. Wang, R.~L. Truby, L.~Chin, and D.~Rus.
\newblock Machine learning best practices for soft robot proprioception.
\newblock In \emph{2023 IEEE/RSJ International Conference on Intelligent Robots and Systems (IROS)}, pages 2564--2571. IEEE, 2023.

\bibitem[Westwood et~al.(2007)]{westwood2007sofa}
J.~Westwood et~al.
\newblock Sofa--an open source framework for medical simulation.
\newblock \emph{Medicine Meets Virtual Reality 15: In Vivo, in Vitro, in Silico: Designing the Next in Medicine}, 125:\penalty0 13, 2007.

\end{thebibliography}

\newpage
\appendix

\section{Task Frames}
\begin{figure}[!ht]
    \centering
    \includegraphics[width=\columnwidth]{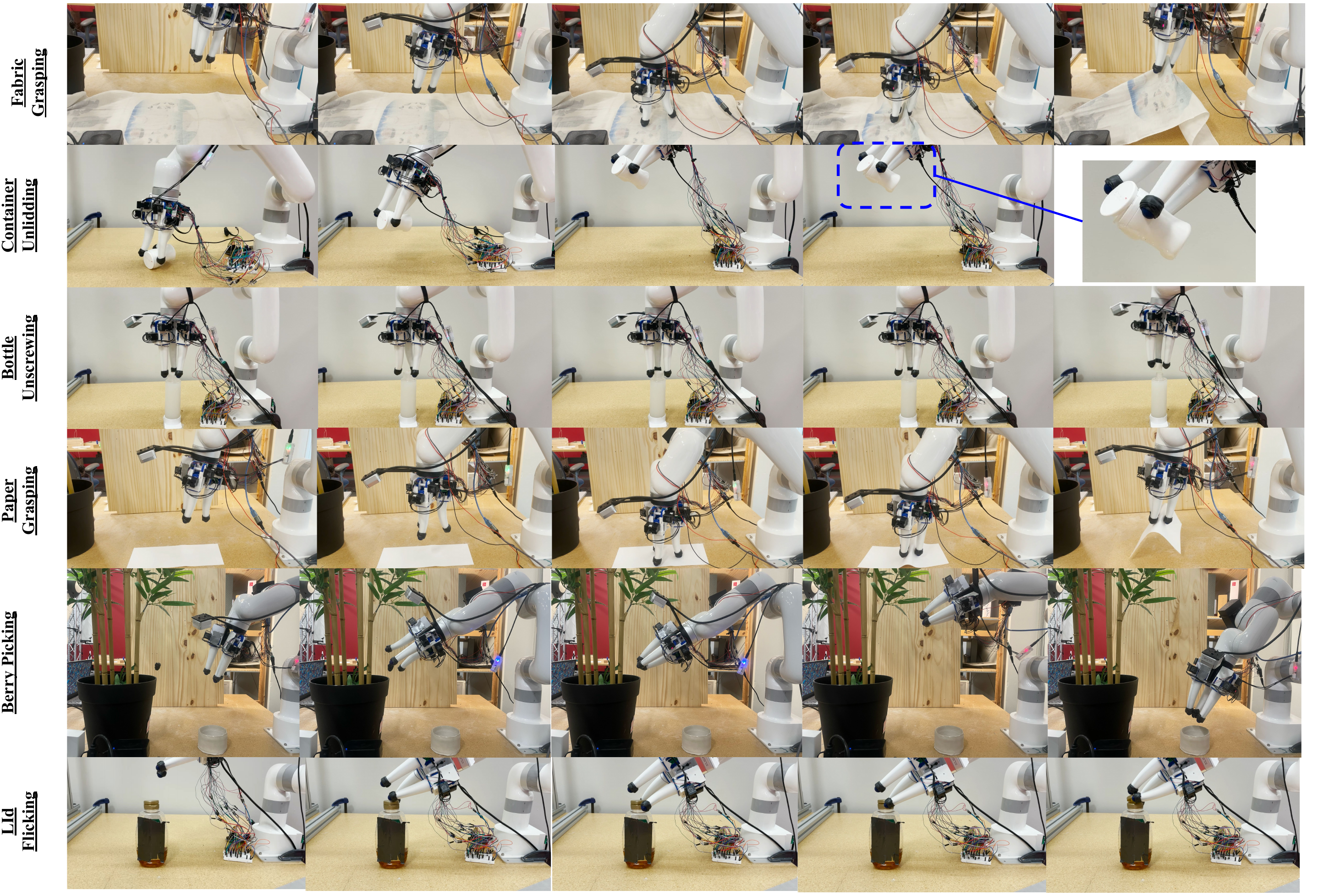} 
    \caption{\textbf{Tasks}}
    \label{fig:app:sensor_signals}
\end{figure}

\section{Shape Estimation and Tracking Evaluation}

\begin{figure}[!ht]
\includegraphics[width=1.\columnwidth]{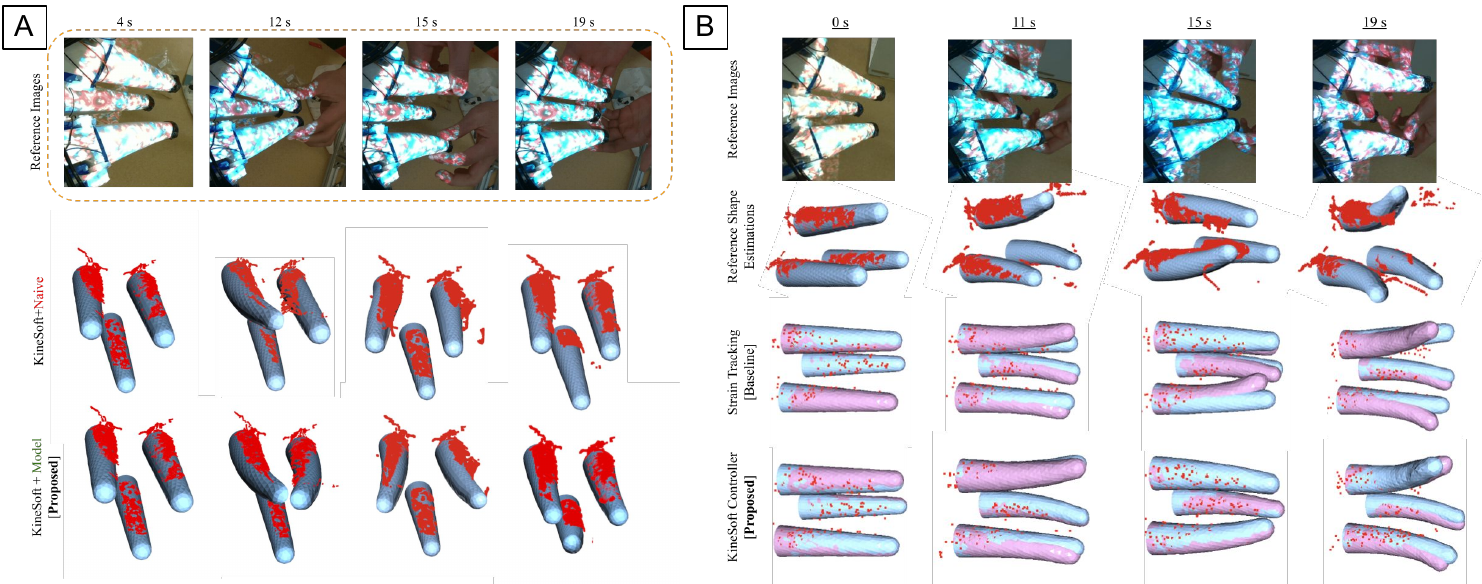}
\caption{\textbf{Shape estimation and trajectory tracking performance evaluation.} We provide each of the shape estimation models and controllers with kinesthetically deformed shape trajectories. \BoxedLabel{A}: Shape estimation model comparisons with real-world ground-truth data (red points). \BoxedLabel{B}: Shape tracking comparisons with the real-world ground-truth data (red points), references shapes (red), and achieved shapes (blue). }
\label{figure:trajectory_tracking_results}
\end{figure}

\section{Details on Sensor Model}
We assume the embedded sensors are perfectly incompressible and isotropic, a common assumption in soft body mechanics for highly elastic rubber, particularly when infused with particle fillers~\citep{starkova2010poisson}. These fillers, like those used in the off-the-shelf conductive rubbers embedded in MOE, enable the sensors to exhibit changes in resistivity when stretched. The sensors have a cylindrical shape, so we model the relationship between the cross-sectional area and the strain in the axial direction for sensor $i$ at time $t \geq 0$ with the incompressibility assumption as:
\begin{equation}
    L_{i, 0}A_{i,0} = L_{i,t} A_{i,t},
    \label{eq:incompress}
\end{equation}
where $L_{i,0}$ and $A_{i,0}$ are the initial length and cross-sectional area; $L_{i,t}$ and $A_{i,t}$ are the corresponding values at time $t$.

For conductive materials, resistance generally has a linear relationship with strain. The observed resistance for the sensor indexed at $i$ is given by:
\begin{equation}
    R_{i,t} = \rho_i \frac{L_{i,t}}{A_{i,t}},
    \label{eq:linear_R}
\end{equation}
\noindent where $\rho_i$ is the conductivity factor, assumed to be constant for sensor $i$ across time. Relating Equation~\ref{eq:incompress} and Equation~\ref{eq:linear_R}, we derive:
\begin{equation}
    \sqrt{\frac{R_{i,t}}{R_{i,0}}}-1 = \frac{L_{i,t} - L_{i,0}}{L_{i,0}}.
    \label{eq:perf_relationship}
\end{equation}
This relationship is independent of the material conductivity $\rho_i$, enabling a direct mapping from observed resistance to strain. However, in real-world applications, fabrication imperfections, such as connecting wires to the DAQ boards, can introduce errors into the initial length of the embedded sensors. These imperfections result in a deviation between the real sensor lengths ($L_{i,0}^R$, $L_{i,t}^R$) and simulated sensor lengths ($L_{i,0}^S$, $L_{i,t}^S$):
\[
L_{i,0}^R = L_{i,0}^S + \epsilon_i, \quad L_{i,t}^R = L_{i,t}^S + \epsilon_i,
\]
\noindent where $\epsilon_i$ is a constant error specific to each sensor $i$. This error propagates to the strain relationship as:
\begin{equation}
    \frac{L_{i,t}^R - L_{i,0}^R}{L_{i,0}^R} = \frac{1}{1 + \frac{\epsilon_i}{L_{i,0}^S}} \cdot \frac{L_{i,t}^S - L_{i,0}^S}{L_{i,0}^S}.
    \label{eq:strain_diff}
\end{equation}

The constant factor $\frac{1}{1 + \frac{\epsilon_i}{L_{i,0}^S}}$ can be denoted as $\kappa_i \in \mathbf{\kappa}$, representing the constant correction factor for sensor $i$. Substituting this into Equation~\ref{eq:perf_relationship}, we obtain:
\begin{equation}
    \sqrt{\frac{R_{i,t}}{R_{i,0}}} - 1 = \kappa_i \frac{L_{i,t}^S - L_{i,0}^S}{L_{i,0}^S},
\end{equation}
where the observed resistances $R_{i,t}, R_{i,0}$ are measured with the DAQ setup. For the $n$ embedded sensors, aligning the simulated and observed distributions involves optimizing the constant correction parameters $\kappa_0, \kappa_1, \dots, \kappa_{n-1}$.


\begin{algorithm}
\caption{Domain Alignment Optimization}\label{alg:domain_align}
\begin{algorithmic}
\Require Shape predictor $f_\theta$, transformations $\{T\}$, observations $\mathcal{P}_\text{obs}$, initial resistances $R_0$
\State Initialize $\theta = \{\mathbf{\kappa} \in \mathbb{R}^{24}, \mathbf{\phi} \in \mathbb{R}^3\} \gets \mathbf{0}$
\State Initialize CMA-ES optimizer $\mathcal{O}(\theta)$
\While{not converged}
    \State Sample candidates $\{\theta\} \sim \mathcal{O}$
    \For{each $\theta$}
        \State $\Delta R \gets \sqrt{R/R_0} - 1$ \Comment{Strain model for resistances}
        \State Apply correction: $S \gets \Delta R \cdot \kappa_{\mathbb{I}[\Delta R < 0]}$
        \State $\mathbf{V} \gets f_\theta(S)$ \Comment{Predict vertices}
        \State $\mathcal{P}_\text{pred} \gets \bigcup \mathbf{V} \cdot T \cdot [R_y(\phi) | \mathbf{0}]$ \Comment{Combine}
        \State $\ell \gets \mathcal{L}_\text{UCD}(\mathcal{P}_\text{obs}, \mathcal{P}_\text{pred})$ \Comment{Compute error}
    \EndFor
    \State Update $\mathcal{O}$ with candidates and losses
\EndWhile
\State \Return $\theta^*$ with minimum loss
\end{algorithmic}
\end{algorithm}

\section{Baselines}
For shape estimation, we compare with analytical and learning-based baselines: 

\parab{Constant curvature model}~\citep{della2020improved,yoo2021analytical}. Constant curvature model is a common representation for the continuum deformation behavior of soft robot that parametrizes the shape with a single curvature curve~\citep{zhang2023machine}. Typically, the independent parameters of the state of the robot are defined by $r_\mathrm{curve}$ and $\theta_\mathrm{curve}$. Assuming a constant length, $L_\mathrm{curve}$ of the robot, we get the constraint: %
\[ L_\mathrm{curve} = r_\mathrm{curve} \theta_\mathrm{curve}.\] %
In typical applications, additional term $\phi_\mathrm{curve}$ is introduced to represent the plane of bending~\citep{katzschmann2019dynamic}. We implemented this simplified representation for soft robot shape using the proposed strain model as outlined in Section~\ref{sec: domain_align} and fitting $r_\mathrm{curve}$ and $\theta_\mathrm{curve}$ to the observed strains in each side of the curve. We transformed the cross-section boundary to the curve during the evaluation and measured the chamfer distance to the reference. 

\parab{DeepSoRo}~\citep{wang2020real}. DeepSoRo architecture deploys a FoldingNet~\citep{yang2018foldingnet} decoder conditioned on visual observations to predict the current shape of a deformable body. Crucially, it is trained with chamfer distance and originally trained on partial real-world shape observations, resulting in partial point cloud reconstruction outputs without frame-to-frame correspondences. Additionally, the model directly outputs the point cloud positions in contrast to KineSoft, which learns a deformation field and produces vertex displacement with frame-to-frame correspondences. We augment DeepSoRo for evaluation by training the model on KineSoft's simulated training data and using the proposed domain alignment process.

\parab{Shape-tracking Baselines.} For shape tracking and task performance evaluation we provide the results against the following: \textbf{Strain Policy}: Strain policy, based on prior works that directly use sensor readings without intermediate representations for learning manipulation policies~\citep{sieler2023dexterous}, uses raw sensor measurements instead of reconstructed shapes. For shape tracking evaluation, we modified the low-level controller from Section~\ref{sec:shape_controller} to track reference sensor readings directly through proportional tendon actuation. For task performance evaluation, we trained a diffusion policy using the same 50 demonstrations we use for KineSoft, but with raw strain signals and wrist-mounted camera observations as input states.

\section{Data Generation and Shape Estimation Model Training}
To train the model, we generate a large dataset of deformed meshes using SOFA (Simulation Open Framework Architecture)~\citep{westwood2007sofa}. We simulate a tetrahedral finite-element mesh of the MOE finger with a Neo-Hookean hyperelastic material model parameterized by elastic material properties that are randomized at runtime.

We model the tendon actuation with massless, inextensible cables running through a series of fixed points within the finger body. We discretize each tendon path to segments defined by 3D attachment points embedded in the tetrahedral mesh. The cable constraint applies forces to these points to maintain constant length while allowing sliding, effectively simulating the mechanical behavior of Bowden cable transmission. The soft body scene is solved with an implicit Euler time integration scheme and uses a conjugate gradient solver for the system matrices. We generate training data by randomly sampling tendon actuation commands within the feasible range and recording the resulting deformed vertex positions and embedded sensor strains. To simulate rich deformation behaviors including contact-like effects, we apply random external forces to the finger surface. These forces are randomly applied over time with sufficiently large radii to ensure smooth deformations that mimic natural contact interactions, without requiring explicit and difficult-to-model contacts in the scene. 

We train the model using a mean squared error (MSE) loss on vertex displacements:
\[
\mathcal{L} = \frac{1}{3N}\sum_{j=1}^3\sum_{i=1}^N \|\Delta \mathbf{v}_{j,i} - \Delta \mathbf{v}_{j,i}^*\|^2,
\]
where $\Delta \mathbf{v}_{j,i}^*$ represents the ground-truth displacement for vertex $i$ of mesh finger $j$. This choice of loss function provides strong supervision by enforcing explicit vertex-wise correspondence between predicted and ground-truth meshes. Because we leverage simulated data to train the model, we can exploit the vertex-level correspondences in the meshes. We contrast this to prior works that relied on chamfer distance loss over real-world partial observations~\citep{wang2020real}, MSE loss ensures that each vertex learns to track its specific local deformation patterns, enabling precise reconstruction of the full finger shape.

\section{Experiment Setup}
\begin{figure}[!ht]
\centering
\includegraphics[width=0.8\columnwidth]{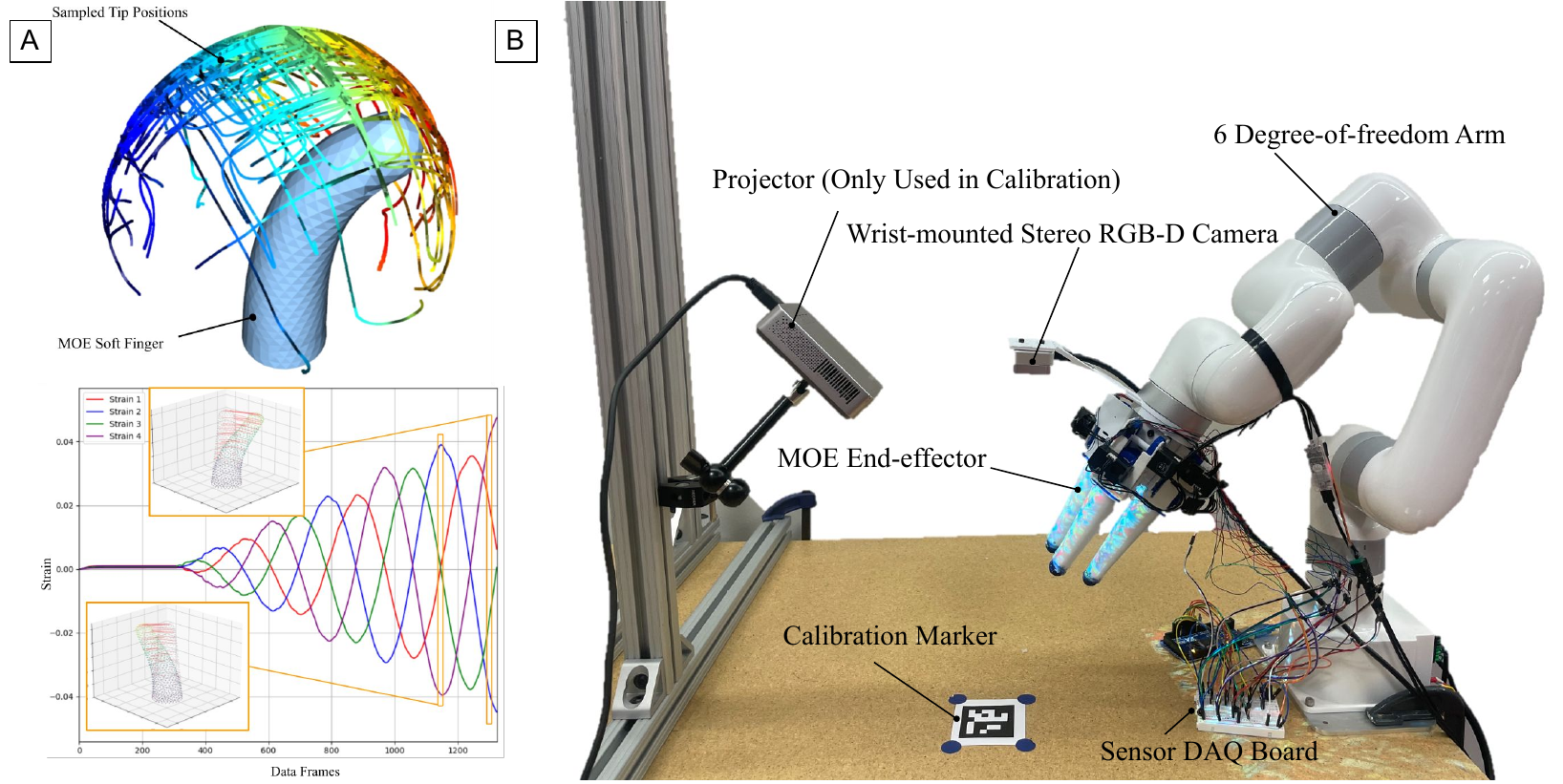}
\caption{\textbf{Simulation and real-world setup}. \BoxedLabel{A}: Simulated robot workspace and sample of simulated strain signals. \BoxedLabel{B}: Real-world robot setup with the projected patterns to improve ground-truth shape observations for evaluation and calibration.}
\label{figure:calibration_setup}
\end{figure}
\newpage
\section{Failiure Cases}
We conducted stress testing through 1,000 continuous servo cycles over 14 hours, observing maximum signal drift of 8.3\% in a sensor (Fig.~\ref{fig:rebuttal}A) that resulted in $0.062 \pm 0.0056$ mm mean vertex error. After rest, the signals returned to baseline levels and our original manipulation experiments were conducted over several weeks without recalibration. We appreciate the reviewers highlighting this important concern and will update the manuscript with these discussions on long-term sensor reliability.

KineSoft failures occurred mainly when fingers lost or made unintended contact during manipulation transitions—a limitation of kinematic imitation without tactile sensing, which we will highlight in the limitations section. Baseline strain policies failed significantly more, as demonstration strain patterns often cannot be directly reproduced through tendon actuation. The results support KineSoft's approach of using shape as an intermediate representation between demonstration and execution modes.
\begin{figure}[!ht]
    \centering
    \includegraphics[width=0.7\columnwidth]{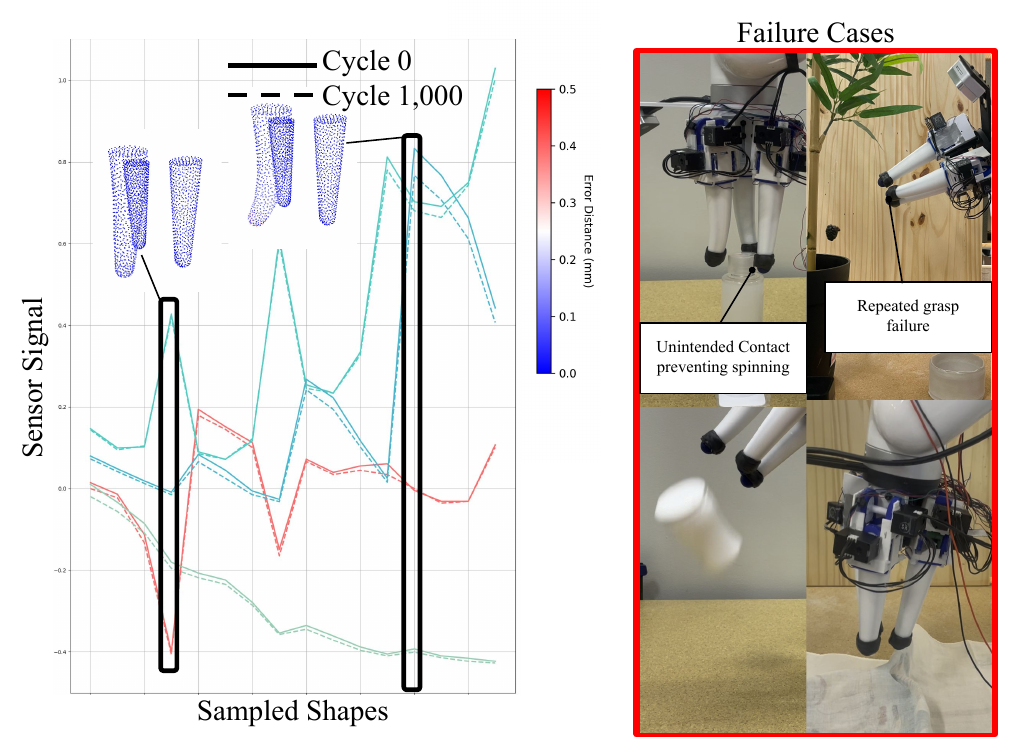} 
    \caption{\textbf{Left:} Sensor signal degradation after 1000 cycles. \textbf{Right:} Failiure cases for manipulation tasks.}
    \label{fig:app:sensor_signals}
\end{figure}

\newpage
\section{Sensor Signals}

\begin{figure}[!ht]
    \centering
    \includegraphics[width=\columnwidth]{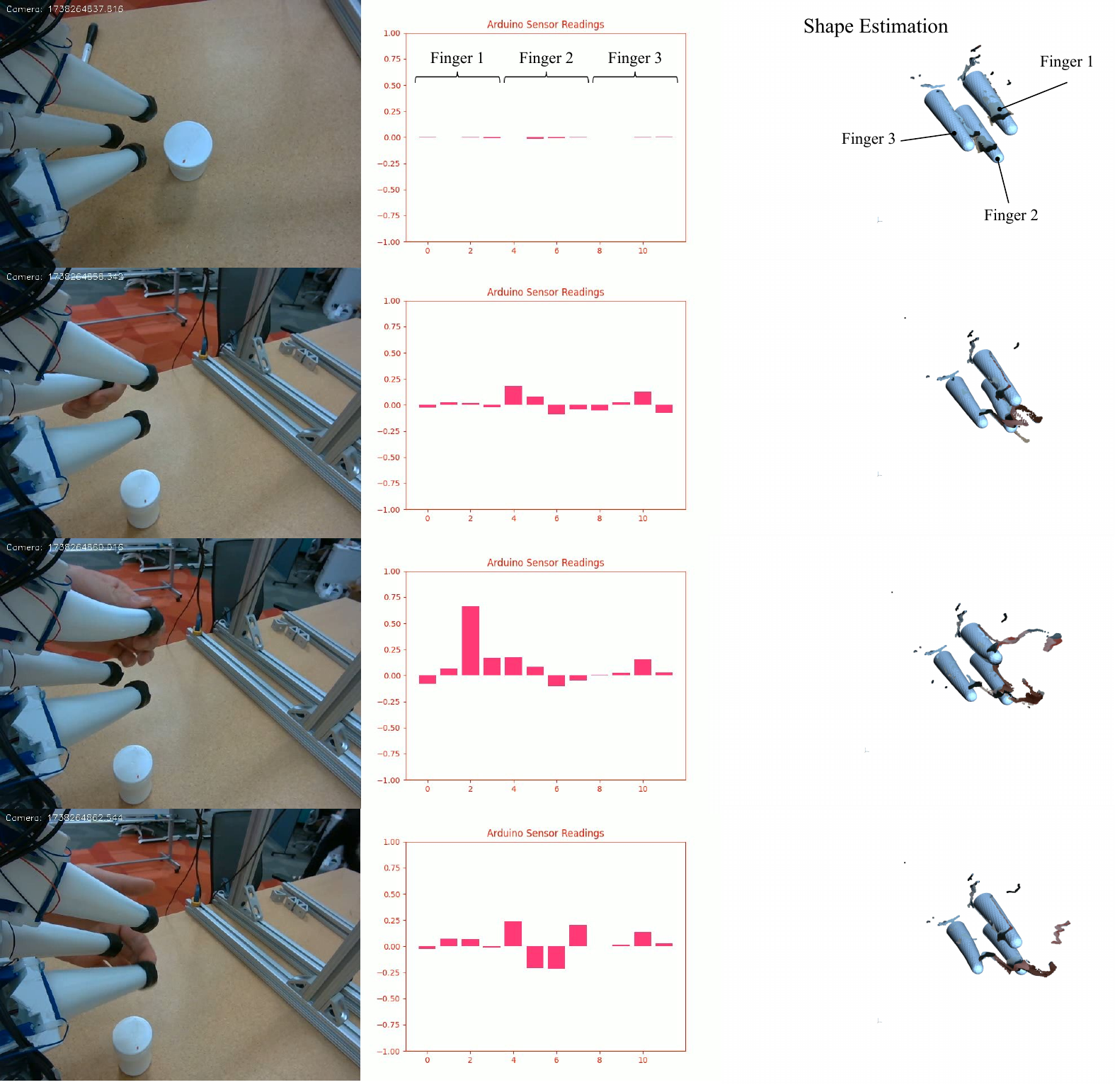} 
    \caption{\textbf{Sensor signals and corresponding shape estimation}}
    \label{fig:app:sensor_signals}
\end{figure}

\end{document}